\documentclass{article}
\pdfoutput=1
\usepackage{ijcai23}

\usepackage{times}
\usepackage{soul}
\usepackage{url}
\usepackage[hidelinks]{hyperref}
\usepackage[utf8]{inputenc}
\usepackage[small]{caption}
\usepackage{graphicx}
\usepackage{amsmath}
\usepackage{amsthm}
\usepackage{booktabs}
\usepackage{algorithm}
\usepackage{algorithmic}
\usepackage[switch]{lineno}

\usepackage{bbding}
\usepackage{amssymb}
\usepackage{dsfont}
\usepackage{pifont}
\usepackage{multirow}
\usepackage{multicol}
\usepackage{subfigure}
\usepackage{hyperref}
\usepackage[T1]{fontenc}
\usepackage{aecompl}
\usepackage{amsmath}
\usepackage{makecell}

\title{A Prototype Multiple-perspective Mining Framework for
Semi-supervised Domain Adaptation}

\author{
Xinyang Huang$^{1,2}$
\and
Chuang Zhu$^1$\thanks{Corresponding Author}
\and
Wenkai Chen$^1$
\affiliations
$^1$School of Artificial Intelligence, Beijing University of Posts and Telecommunications\\
$^2$School of Artificial Intelligence, Xidian University\\
\emails
hsinyanghuang7@gmail.com,
\{czhu, wkchen\}@bupt.edu.com
}

\let\oldalign\align
\let\oldendalign\endalign

\renewenvironment{align}
  {\linenomathNonumbers\oldalign}
  {\oldendalign\endlinenomath}

\let\oldequation\equation
\let\oldendequation\endequation

\renewenvironment{equation}
  {\linenomathNonumbers\oldequation}
  {\oldendequation\endlinenomath}

\begin{document}

\maketitle

\begin{abstract}
In semi-supervised domain adaptation (SSDA), a few labeled target samples of each class help the model to transfer knowledge representation from the fully labeled source domain to the target domain. The benefits of using prototypes generated by these additional labels were ignored by existing methods. To fill this gap, we propose a novel Prototype Multiple-perspective Mining (ProMM) framework to better tap the potential of target prototypes. We first introduce a prototype pseudo-label aggregation based on optimal transport to help the model generate robust prototypes using target labels and align the feature distribution of unlabeled target samples with the prototype. For cross-domain knowledge transfer, we propose a prototype alignment loss to help the model align source samples and target prototypes from the same categories. In addition to considering the relationship between samples and prototypes, we further propose a dual consistency based on prototype similarity and linear classifier to promote the learning of discriminative target features between different samples from the same batch. Extensive experiments on three datasets, including DomainNet, VisDA2017, and Office-Home, demonstrate that our proposed method achieves state-of-the-art performance in SSDA. Our code will be available soon.
\end{abstract}

\section{Introduction}

The deep neural network has achieved great success in various visual tasks. 
However, due to the high cost of labeled data it requires and the degradation of model performance when deploying models in a new domain (target), unsupervised domain adaptation (UDA) has been proposed. To further improve the model performance, many semi-supervised domain adaptation (SSDA) works are proposed by adding a few (e.g., one sample per class) labeled target samples based on UDA.
Compared to UDA, the key to SSDA is whether the model can make better use of the additional limited target samples to help the model better fit the learned features from the source domain to the target domain.

\begin{figure}[t]
\centering
\includegraphics[scale=0.25]{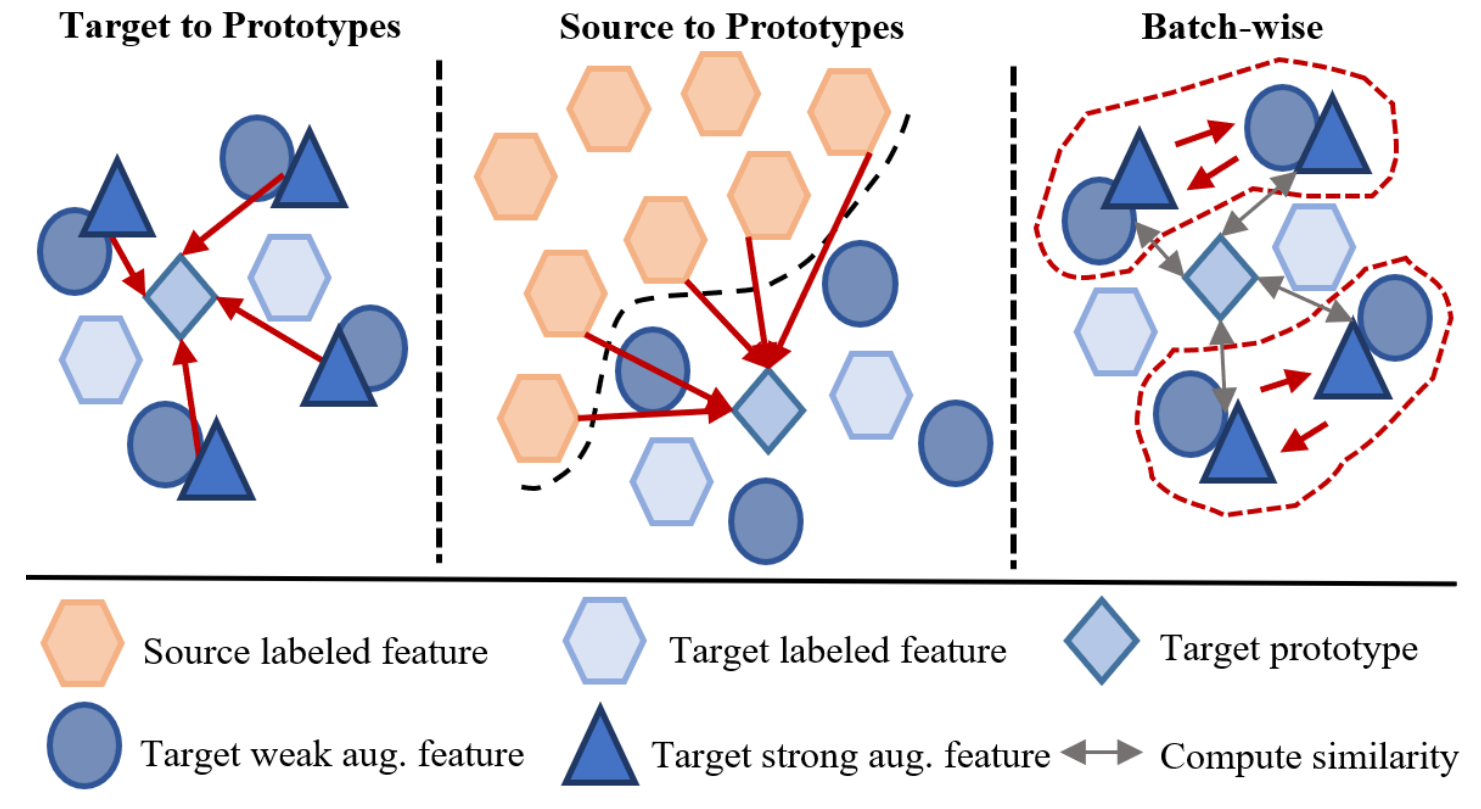}
\caption{Overview of our Prototype Multiple-perspective Mining (ProMM) framework. 
Our ProMM framework helps models transfer knowledge using the target prototype from (1) intra-domain level, (2) inter-domain level, and (3) batch level.
The arrows represent the feature alignment, and \textit{aug.} represents augment.}
\label{figure1}
\end{figure}

In recent years, to help model knowledge transfer, some existing SSDA studies make better use of unlabeled samples in target domains by combining self-supervised learning \cite{perez20222} and semi-supervised learning \cite{li2021cross,yan2022multi}.
However, they ignore the potential of labeled samples in the target domain.
Although \cite{li2021ecacl,singh2021clda} use few target samples by building a prototype, they only use the prototype from one level unilaterally and ignore the additional knowledge learned from other levels (e.g. intra-domain level). 
At the same time, it is not robust to use only a very limited number of label or pseudo-label samples to build prototypes.
To fill this gap, we propose a robust Prototype Multiple-perspective Mining  (ProMM) framework. 
Our prototype is updated by both labeled and unlabeled target samples which helps the model to transfer knowledge across domains more robustly.
Our ProMM makes full use of the target prototype from three levels: intra-domain, inter-domain, and batch-wise to help the model better transfer knowledge through target samples, as shown in Figure~\ref{figure1}.

Specifically, at the intra-domain level, we propose a pseudo-labeling strategy based on the optimal transport to help the model align the data distribution of labeled and unlabeled samples in the target domain, obtaining a more compact intra-domain feature distribution and robust pseudo-labels.
This method improves the pseudo-labels of target samples by resolving the optimal transport plan between the weakly augmented target samples and the target prototype.
At the same time, the transport plan is also applied to the strongly augment view to form a consistency constraint.
At the inter-domain level, we cross-align the source samples with the same class of the target prototype.
Inter-domain prototype alignment helps the model learn better about cross-domain knowledge transfer and category alignment.
At the batch level, different from work \cite{yan2022multi}, we consider a mini-batch of samples to calculate class correlation matrices between predictions with different augments from two perspectives of prediction probability and prototype similarity, increasing the correlation in the same class and reducing the correlation from different classes.
Classifiers from different perspectives represent features comprehensive to learn the relationship between the batch-wise target samples.
Combining the above three, through our ProMM framework, the model can capture the knowledge of target domain samples from different levels, to better enable the model to learn more comprehensive and complementary domain-adaptive knowledge.

Our main contributions can be summarized as follows:

(1) We propose a novel Prototype Multiple-perspective Mining (ProMM) framework, which exploits the potential of labeled target samples by making full use of the target prototype from multiple levels.

(2) We propose a pseudo-label aggregation based on intra-domain optimal transport to SSDA, which helps the model form a more compact target domain robustly.

(3) We propose a batch-wise dual consistency, which helps the model learn more distinctive target representations from different perspectives.

(4) Experiments have shown that our ProMM implements a new state-of-the-art result in most SSDA problems.

\section{Related Works}
\textbf{Unsupervised domain adaptation.} Unsupervised domain adaptation (UDA) \cite{pan2010domain} has made some exciting achievements. 
It aims to transfer the knowledge learned from the labeled source domain to the unlabeled target domain.
The methods based on feature space alignment \cite{long2017deep,sun2016deep} usually use joint distribution to make the two domains as close as possible in the feature space, to reduce the differences between the two domains.
Some methods based on GAN \cite{chen2019joint,long2018conditional,pei2018multi} are also popular.
They calibrate the distribution of the source domain and target domain by generating codes that cannot be distinguished from the perspective of the discriminator, which is trained to classify the target domain.
Despite significant advances in UDA, UDA methods do not perform well in SSDA \cite{saito2019semi}, which is our main reason for focusing on SSDA.

\noindent\textbf{Semi-supervised domain adaptation.} In SSDA, the problem assumes that there are a few labeled samples in the target domain.
It can be thought of as a combination of semi-supervised learning (SSL) and DA. 
Some methods \cite{saito2019semi,li2021cross} propose using Adversarial training to adjust source and target distributions, clustering similar target samples together through different clustering strategies, and separating different samples to reduce intra-domain gaps.
Some methods \cite{li2021ecacl,singh2021clda} use the idea of self-supervised learning to build target prototypes and help model transformation across domains by comparing learning methods.
They only build the prototype with very limited labels or high-noise pseudo-labels, ignoring the importance of different sample relationships.
The methods based on SSL \cite{perez20222} use different losses to enhance the consistency between the feature representations of unlabeled samples and
\cite{yan2022multi} standardizes the consistency of different views of target domain samples at three levels, which facilitates learning more representative target features from each other.
However, they all ignore the importance of making further use of the labeled target samples.
In this work, we use the prototype to tap the potential of labeled target samples from multiple levels to help the model learn comprehensive and complementary feature representations.
\section{Methodology}
\begin{figure*}[t]
\centering
\includegraphics[scale=0.32]{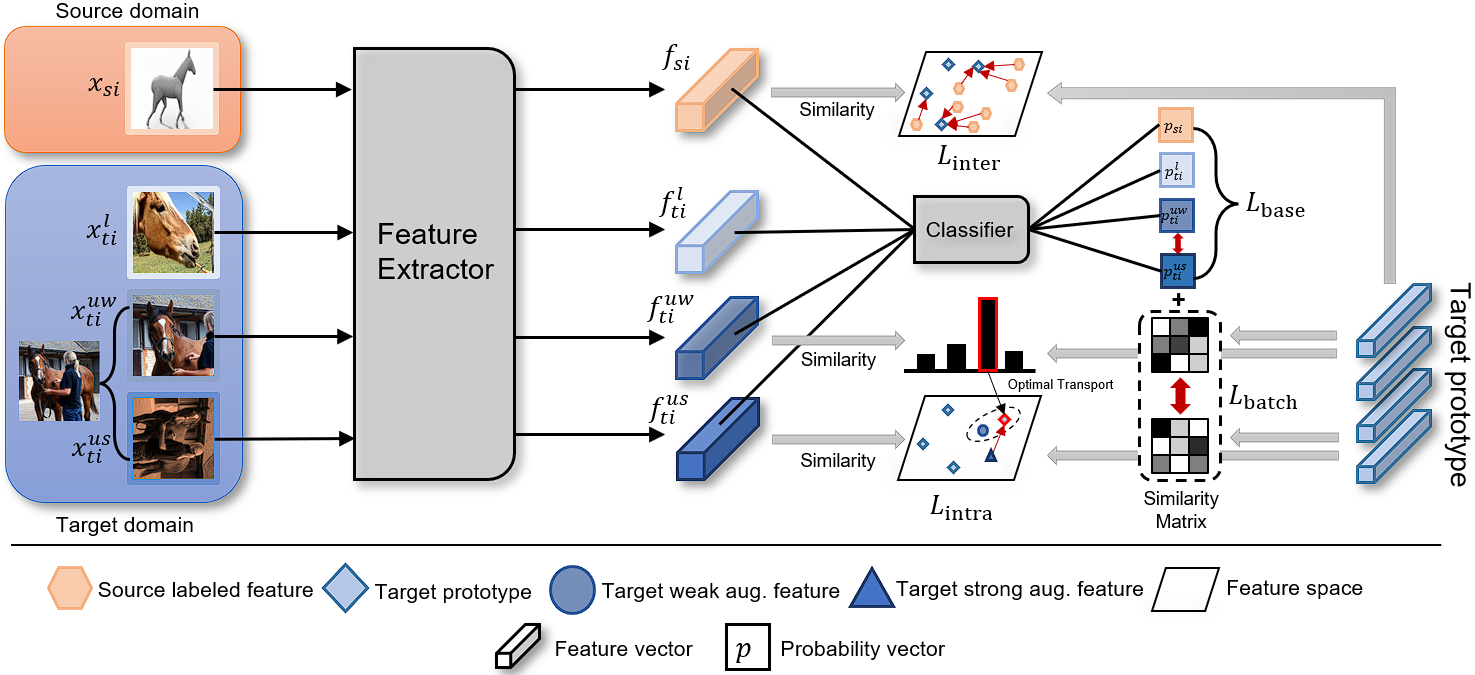}
\caption{The structure of our ProMM framework.
First, the target samples are weakly and strongly augmented and then pass through the classifier together with the source samples to calculate the base loss.
For the intra-domain level, the weakly augmented target samples generate pseudo-labels with the optimal transfer plan computed with the target prototype and compute the consistency loss with the strongly augmented samples.
For the inter-domain level, the similarity loss between source samples and the target prototype of corresponding categories is computed to achieve cross-domain knowledge transfer.
Finally, the dual consistency loss of the two augmented views in each mini-batch is considered from the perspective of linear and prototype-based classifiers.}
\label{figure2}
\end{figure*}
In this section, we first specify the definition and notation of SSDA and then introduce the proposed Prototype Multiple-perspective Mining (ProMM) framework from three levels.
\subsection{Framework}
In SSDA, we sample datasets from two different distributions. 
In this setting, we can access labeled source samples $\mathcal{D}_{s}=\left\{\left(x_{si}, y_{si}\right)\right\}_{i=1}^{N_{s}}$ sampled from the source distribution $P_s(X,Y)$. 
We also have a limited number of labeled target samples $\mathcal{D}_{t}^l=\left\{\left(x_{ti}^l, y_{ti}^l\right)\right\}_{i=1}^{N_{t}^l}$ and a large number of unlabeled samples $\mathcal{D}_{t}^u=\left\{\left(x_{ti}^u\right)\right\}_{i=1}^{N_{t}^u}$ both from the target distribution $P_t(X, Y)$. 
This two distribution satisfy: $P_s(Y)=P_t(Y)$ and $P_s(X|Y)\neq P_t(X|Y)$.
The framework is composed of a feature extractor $G$ and a linear classifier $F$.
An outline of our ProMM framework is illustrated in Figure~\ref{figure2}.

Following \cite{sohn2020fixmatch}, we feed the labeled source samples and the labeled target samples into $G$ to obtain their feature representations $f_{s}$, $f_{t}^l\in \mathbb{R}^d$, and then get the probability prediction $p_{s}$, $p_{t}^l\in \mathbb{R}^c$ through $F$.
Similarly, we generate two different views, for each unlabeled target sample $x_{ti}^u$ by weak and strong augment, represented as $x_{ti}^{uw}$ and $x_{ti}^{us}$.
Then the target samples of these two views are fed to the same feature extractor $G$ to generate representations $f_{ti}^{uw}$ and $f_{ti}^{us}$.
Finally, the probability prediction $p_{ti}^{uw}$, $p_{ti}^{us}$ are obtained through the same classifier $F$.
We can calculate the standard cross-entropy loss by using labeled sample pairs from two domains and consider a simple pseudo-label regularization with different views of unlabeled samples.
Specifically, we optimize the following baseline:
\begin{align}
L_{\text {base}} & = -\left(\sum_{x\in x_s \cup x_t^l} y \log p+\sum_{x\in x_t^u} \mathds{1}_{\{p_t^{uw} \geq \tau_1\}} \log p_t^{us}\right),
\end{align}
where $\mathds{1}$ is an indicator function, $y$ is the label of labeled samples $x$, and $\tau_1$ is the pseudo-label threshold of the regularization.
In the following, we will further introduce our framework from three different levels.
\subsection{Intra-domain Pseudo-label Aggregation}
A compact target domain can help the model make better use of pseudo-labels and cross-domain knowledge transfer. 
We propose a novel pseudo-label aggregation strategy to align the unlabeled and labeled target data robustly and accurately.

\begin{figure}[t]
\centering
\includegraphics[scale=0.26]{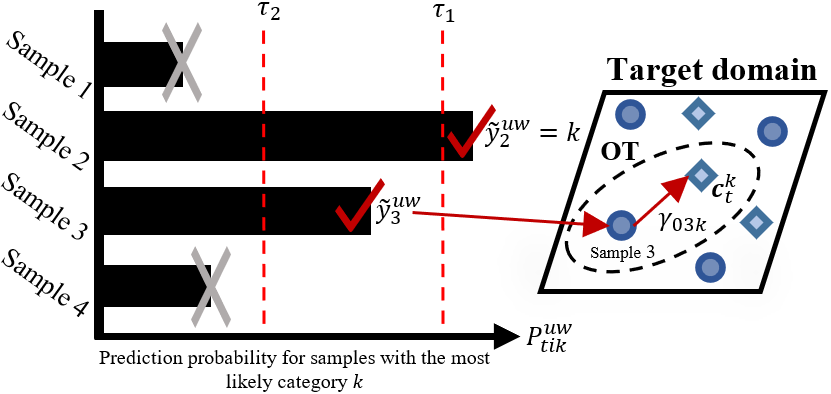}
\caption{: Illustration of the pseudo-label strategy. For the
sample with a relatively high probability of predicting a
maximum class (Sample 2 in the figure), we directly output
the category corresponding to that probability as its pseudo-label. For this sample (Sample 3), which has a relatively low
prediction probability, we use its optimized transport as its
pseudo-label, which is more accurate than its prediction probability. For other samples (Samples 1 and 4), we do not give
them pseudo-labels.}
\label{number}
\end{figure}

To make full use of the labeled target samples, we initialize the target prototype:
\begin{align}
\mathbf{c}^{k}_{t} = \frac{\sum_{i = 1}^{i = N_t^l} \mathds{1}_{\left\{y_{ti}^{l} = k\right\}} f_{ti}^l}{\sum_{i = 1}^{i = N_t^l} \mathds{1}_{\left\{y_{ti}^{l} = k\right\}}},
\end{align}
where $y_{ti}^{l}$ is the label of the $i$-th labeled target sample.
The optimal transport (OT) is used to align the inter-domain level representation for domain adaptation \cite{courty2017joint,yan2022multi}. 
It finds the optimal coupling plan $\gamma_0$ according to the given transport function to minimize the total transport cost.
To make better use of the additional data provided by SSDA, we propose to apply OT to the intra-domain feature aggregation instead of the inter-domain.
For labeled and weakly augment unlabeled samples, we assume that:
\begin{align}
\gamma_0 & = \underset{\gamma \in \mathcal{B}}{\arg \min }\left\langle\gamma, \mathbf{C}^{w}\right\rangle_{F},
\end{align}
\begin{align}
\mathcal{B} = \left\{\gamma \in\left(\mathbb{R}^{+}\right)^{N_{t}^l \times N_{t}^u} \mid \gamma \mathbf{1}_{N_{t}^l} = \mu_{t}^l, \gamma^{\top} \mathbf{1}_{N_{t}^u} = \mu_{t}^u\right\},
\end{align}
where $\gamma_{0i,j}$ means the transport plan between the $i$-th labeled sample and the $j$-th unlabeled sample in the target domain, $\langle\cdot, \cdot\rangle_F$ represents Frobenius inner product, $\mathbf{1}_d$ is 1s' $d$-dimensional vector, $\mu_t^l \in \mathbb{R}^{N_t^l}$, $\mu_t^u \in \mathbb{R}^{N_t^u}$ are the empirical distributions of labeled and unlabeled samples respectively, and the default is a uniform distribution.
$\mathbf{C}^w\in \mathbb{R}^{N_t^l \times N_t^u}$ is a cost matrix for indicating each transportation.
In consideration of the robustness and accuracy of the transportation, we propose to use the target prototype to replace the labeled samples in the transport plan. 
This not only makes the feature representation more representative but also alleviates the negative alignment caused by the lack of some classes in the minibatch of samples.
The cost matrix is as follows:
\begin{align}\label{sim_w_prototype}
\mathbf{C}_{i, j}^{w} & = 1-\mathbf{c}_{t}^{i\top} f_{tj}^{uw},
\end{align}
where each element in the cost matrix $\mathbf{C}^w_{i, j}$ represents the degree of difference between the $i$-th target prototype $\mathbf{c}_{t}^{i}$ and the $j$-th weak augment target feature $f^{uw}_{tj}$.
The resulting OT plan $\gamma_0$ helps the model better align the distribution of different samples in the target domain.

To further improve the robustness of the prototype, we add unlabeled target samples to the update of the prototype and regard the OT plan as a supplement to the pseudo-label strategy. 
Specifically, we consider the following pseudo-label strategy:
\begin{equation}
\begin{aligned}
\tilde y^{uw}_i = \left\{\begin{array}{ll}
argmax(\tilde{p}^{uw}_{ti}), & \text { if }\tilde{p}^{uw}_{ti}\geq\tau_1  \\
\gamma_{0i}, & \text { if } \tilde{p}^{uw}_{ti} <\tau_1 \text { and } \tilde{p}^{uw}_{ti}\geq\tau_2,\\
0, & \text { otherwise }
\end{array}\right.
\end{aligned}
\label{label}
\end{equation}
where $\tau_2$ is the confidence threshold for using a pseudo-label supplementation strategy, $\tilde{p}_{ti}^{uw}$is the maximum prediction probability of the $i$-th unlabeled weak sample in all categories, and $argmax(\cdot)$ represents the category corresponding to the prediction probability.
When the distribution of unlabeled data is close to that of labeled data, this module can better use the prototype initialized by labeled data to assign pseudo-labels to unlabeled data through the proposed strategy.
We make better use of labeled data to improve the robustness of prototypes through the pseudo-label strategy.

For each minibatch, we calculate the feature average of labeled and pseudo-labeled target samples for clustering, and then use the exponential moving average to update the target prototype during training:
\begin{align}
\mathbf{c}_{t}^{k} & = \alpha \mathbf{c}_{t}^{k}+(1-\alpha)\tilde{\mathbf{c}}_{t}^{k},
\end{align}
where $\tilde{\mathbf{c}}_{t}^{k}$ is the target prototype clustered by target samples with labels and pseudo-labels in this minibatch.
Under the guidance of the optimal coupling $\gamma_0$, the feature representation of each strongly augment view target sample can have a consistent mapping plan with the weak augment view, forming a consistency constraint:
\begin{align}
L_{\text {intra}} & = \left\langle\gamma_0, \mathbf{C}^{s}\right\rangle_{F},
\end{align}
where $\mathbf{C}^{s}$ is the similarity matrix between the strong augment target sample and the target prototype similar to Equation \ref{sim_w_prototype}.
As mentioned above, the model can effectively align the distribution of the target prototype and unlabeled samples and pay more attention to the details of the intra-domain level.
\subsection{Inter-domain Alignment}
Knowledge transfer is an essential capability of models in SSDA. 
For the cross-domain level, we can use the relationship between the target prototype and the source sample to naturally transfer the knowledge of the model in the source domain to the target domain. 

Specifically, for each category of target prototype, we can calculate the softmax of the similarity of the corresponding target prototype in the feature space according to the category of the source sample:
\begin{align}
\label{sim_pro}
\mathbf{s}^{k}_{i}=\frac{\exp \left(\textrm{sim}\left(f_{si}, \mathbf{c}^k_{t}\right) / T_1\right)}{\sum_{i=1}^{N_s} \exp \left(\textrm{sim}\left(f_{si}, \mathbf{c}^k_{t}\right) / T_1\right)},
\end{align}
where $\textrm{sim}(\cdot,\cdot)$ means cosine similarity, $f_{si}$ is the feature of i-th source samples, $T_1$ is a scale temperature, and $N_s$ is the number of source samples.
Then, we can calculate the cross-domain prototype alignment loss for each source sample:
\begin{align}
L_{\text {inter}} =-\sum_{k=1}^{C}\sum_{i=1}^{N_s}\mathds{1}\left(y_s= k\right)\log \mathbf{s}^{k}_{i},
\end{align}
where $C$ is the number of classes.
Our prototype is not only composed of labeled targets but also adopts a more robust pseudo-label update strategy to alleviate negative transfer caused by the dispersion of target samples.
\subsection{Batch-wise Dual Consistency}
To make the model more comprehensively learn the representation in the target domain, different from \cite{yan2022multi}, we consider the dual relationship between target features at the batch level.
We increase the confidence difference between different views by sharpening the confidence and then model the clustering of each class as the classification confidence and prototype similarity for all samples of that class in the batch.
Maintain the consistency of the strong and weak views of the class allocation as the positive pair in the same batch, and reduce the similarity between different classes as the negative pair, instead of the sample-wise contrastive learning.

Given the prediction $\mathbf{P}^{uw}\in \mathbb{R}^{N_t\times C}$ of the target sample of a batch, we use the sharpening function to reduce the entropy of label distribution, widening the gap between different confidence, and enhancing the contrast between different views:
\begin{align}
\hat{p}_{i}^{uw} = \frac{p_{i}^{uw \frac{1}{T_2}}}{\sum_{j = 1}^{C} p_{j}^{uw \frac{1}{T_2}}},
\end{align}
where $T_2$ is a temperature hyperparameter, $C$ is the number of classes.
However, it is one-sided to optimize only from the perspective of linear classifiers.
Due to the existence of the target prototype, we propose that batch level relationships be learned dually not only from the perspective of linear classifier but also from the perspective of prototype similarity.
Since linear classifier can assign learnable parameters to each class, while prototype-based classifier only relies on excellent feature representation, we calculate the cross-correlation matrix of strong and weak views from the perspective of linear classifier and prototype-based classifier respectively:
\begin{equation}
\begin{aligned}
\mathbf{R}_l^{ws} &= \mathbf{\hat{P}}^{uw\top}\mathbf{\hat{P}}^{us}\\
\mathbf{R}_p^{ws} &= \mathbf{S}^{uw\top}\mathbf{S}^{us}
,
\end{aligned}
\end{equation}
where $\mathbf{\hat{P}}^{uw}$, $\mathbf{\hat{P}}^{us}$ are sharpening batch probability matrices and $\mathbf{S}^{uw}$, $\mathbf{S}^{us}$ are batch similarity matrices with the prototype from weak and strong views similar to Equation \ref{sim_pro}.
$\mathbf{R}_l^{ws}$, $\mathbf{R}_p^{ws}$ are asymmetric matrices, and each element represents two similarities of different views at the batch level. 
From this, the dual contrast loss of batch class can be calculated:
\begin{equation}
\begin{aligned}
L_{\text {batch}} & = \frac{1}{2 C}\underbrace{(\|\phi(\mathbf{R}_l^{ws})-\mathbf{I}\|_{1}+\left\|\phi\left(\mathbf{R}_{l}^{ws\top}\right)-\mathbf{I}\right\|_{1}}_\text{linear classifier}+\\
&\underbrace{\|\phi(\mathbf{R}_p^{ws})-\mathbf{I}\|_{1}+\left\|\phi\left(\mathbf{R}_{p}^{ws\top}\right)-\mathbf{I}\right\|_{1})}_\text{prototype-based classifier},
\end{aligned}
\end{equation}
where $\phi(\cdot)$ is a normalized function that keeps the row total as 1. 
$\mathbf{I}\in \mathbb{R}^{C \times C}$ is the identity matrix, and $| | \cdot| |_1$ represents the sum of the absolute values of the matrix.

Dual consistency can benefit the framework in learning cross-domain knowledge from two perspectives.
Linear classifiers can use learnable parameters to focus more on the distinguishing dimensions represented by features while suppressing unrelated feature dimensions by assigning higher or lower weights to different dimensions.
In contrast, prototype-based classifiers cannot take advantage of this which requires more discriminative feature representation \cite{xu2022semi}.

At the same time, due to the setting of DA, the prediction of the linear classifier may be biased toward the source domain. 
Comparing with target prototype similarity can alleviate the inaccurate contrast caused by over-fitting the source domain.
We combine the two to encourage the model to learn a more discriminative and accurate relationship between batch level objectives from different perspectives.

\begin{table*}[t]
\centering
\renewcommand\arraystretch{0.9}
\tabcolsep=1.2pt\scalebox{1}{
\begin{tabular}{c|cccccccccccccc | cc}
        \toprule
\multirow{2}{*}{\textbf{Method}} & \multicolumn{2}{c}{R$\rightarrow$C} & \multicolumn{2}{c}{R$\rightarrow$P} & \multicolumn{2}{c}{P$\rightarrow$  C} & \multicolumn{2}{c}{C$\rightarrow$S} & \multicolumn{2}{c}{S$\rightarrow$P} & \multicolumn{2}{c}{R$\rightarrow$S} & \multicolumn{2}{c|}{P$\rightarrow$R} & \multicolumn{2}{c}{Mean}  \\
   & 1-shot & 3-shot         & 1 -shot & 3-shot        & 1 -shot & 3-shot         & 1-shot & 3-shot        & 1-shot & 3-shot           & 1 -shot & 3-shot        & 1-shot & 3-shot         & 1 -shot & 3-shot          \\
                      \midrule
S+T & 55.6   & 60.0  & 60.6 & 62.2 & 56.8 & 59.4 & 50.8 & 55.0   & 56.0   & 59.5  & 46.3 & 50.1  & 71.8   & 73.9  & 56.9 & 60.0   \\
DANN    & 58.2   & 59.8           & 61.4    & 62.8          & 56.3    & 59.6           & 52.8   & 55.4          & 57.4   & 59.9             & 52.2    & 54.9          & 70.3   & 72.2           & 58.4  & 60.7            \\
ENT     & 65.2   & 71.0           & 65.9    & 69.2          & 65.4    & 71.1           & 54.6   & 60.0          & 59.7   & 62.1             & 52.1    & 61.1          & 75.0   & 78.6           & 62.6    & 67.6            \\
MME   & 70.0   & 72.2           & 67.7    & 69.7          & 69.0    & 71.7           & 56.3   & 61.8          & 64.8   & 66.8             & 61.0    & 61.9          & 76.1   & 78.5           & 66.4    & 68.9            \\
BiAT  & 73.0   & 74.9           & 68.0    & 68.8          & 71.6    & 74.6           & 57.9   & 61.5          & 63.9   & 67.5             & 58.5    & 62.1          & 77.0   & 78.6           & 67.1    & 69.7            \\
APE  & 70.4   & 76.6           & 70.8    & 72.1          & 72.9    & 76.7           & 56.7   & 63.1          & 64.5   & 66.1             & 63.0    & 67.8          & 76.6   & 79.4           & 67.6    & 71.7            \\
Con$^2$DA     & 71.3   & 74.2           & 71.8    & 72.1          & 71.1    & 75.0           & 60.0   & 65.7          & 63.5   & 67.1             & 65.2    & 67.1          & 75.7   & 78.6           & 68.4   & 71.4       \\
CDAC  & 77.4   & 79.6           & 74.2    & 75.1          & 75.5    & 79.3           & 67.6   & 69.9          & 71.0   & 73.4             & 69.2    & 72.5          & 80.4   & 81.9           & 73.6    & 76.0            \\
DECOTA    & \textbf{79.1}   & \textbf{80.4}           & 74.9    & 75.2          & 76.9    & 78.7           & 65.1   & 68.6          & 72.0   & 72.7             & 69.7    & 71.9          & 79.6   & 81.5           & 73.9    & 75.6            \\
CLDA    & 76.1   & 77.7           & 75.1    & 75.7          & 71.0    & 76.4           & 63.7   & 69.7          & 70.2   & 73.7             & 67.1    & 71.1          & 80.1   & 82.9           & 71.9    & 75.3       \\    
ECACL    & 75.3   & 79.0           & 74.1    & \textbf{77.3}          & 75.3    & \textbf{79.4}           & 65.0   & 70.6          & 72.1   & 74.6             & 68.1    & 71.6          & 79.7   & 82.4           & 72.8    & 76.4       \\    
MCL   & 77.4   & 79.4           & 74.6    & 76.3          & 75.5    & 78.8           & 66.4   & 70.9          & 74.0   & 74.7             & 70.7    & 72.3          & 82.0   & 83.3           & 74.4    & 76.5       \\    
\midrule
\textbf{ProMM}    & 78.5   & 80.2           & \textbf{75.4}    & 76.5          & \textbf{77.8}    & 78.9           & \textbf{70.2}   & \textbf{72.0}          & \textbf{74.1}   & \textbf{75.4}             & \textbf{72.4}    & \textbf{73.5}         & \textbf{84.0}   & \textbf{84.8}           & \textbf{76.1}    & \textbf{77.4} \\
  \bottomrule
\end{tabular}}
    \caption{ Accuracy (\%) on \textit{\textit{DomainNet}} under the settings of 1-shot and 3-shot using ResNet34 as backbone networks.}
    \label{multi}
\end{table*}

\begin{table}[h]
    \centering
    \renewcommand\arraystretch{0.92}\scalebox{0.9}{
    \begin{tabular}{c|cc}
        \toprule
        \textbf{Method}  & 1-shot & 3-shot \\
        \midrule
        S+T     & 60.2          & 64.6        \\
        ENT  & 63.6          & 72.7        \\
        MME & 68.7          & 70.9        \\
        APE & 78.9          & 81.0        \\
        CDAC & 69.9          & 80.6        \\
        DECOTA & 64.9          & 80.7        \\
        ECACL & 81.1          & 83.3        \\
        MCL & 86.3          & 87.3        \\
       \midrule
     \textbf{ProMM} & \textbf{87.6}          & \textbf{88.4}        \\
        \bottomrule
    \end{tabular}}
    \caption{Mean Class-wise Accuracy (MCA)(\%) on \textit{VisDA2017} using ResNet34 as the backbone network.
}
    \label{visda2017}
\end{table}

\begin{table*}[h]
\centering
\renewcommand\arraystretch{0.9}
\tabcolsep=4.3pt\scalebox{1}{
\begin{tabular}{c|cccccccccccc|c}
\toprule
\textbf{Method}    & R$\rightarrow$ C                  & R$\rightarrow$P                  & R $\rightarrow$A                 & P $\rightarrow$R                  & P $\rightarrow$C                  & P$\rightarrow$ A                  & A$\rightarrow$P                  & A$\rightarrow$C                  & A$\rightarrow$ R                 & C $\rightarrow$ R                   & C $\rightarrow$ A                  & C $\rightarrow$ P                   & Mean                  \\ 
\midrule
\multicolumn{14}{c}{\textbf{1-shot}}\\ 
\midrule
S+T                   & 52.1                 & 78.6                 & 66.2                 & 74.4                 & 48.3                 & 57.2                 & 69.8                 & 50.9                 & 73.8                 & 70.0                 & 56.3                 & 68.1                  & 63.8                  \\
DANN                 & 53.1                 & 74.8                 & 64.5                 & 68.4                 & 51.9                 & 55.7                 & 67.9                 & 52.3                 & 73.9                 & 69.2                 & 54.1                 & 66.8                  & 62.7                  \\
ENT                   & 53.6                 & 81.9                 & 70.4                 & 79.9                 & 51.9                 & 63.0                 & 75.0                 & 52.9                 & 76.7                 & 73.2                 & 63.2                 & 73.6                  & 67.9                  \\
MME                  & 61.9                 & 82.8                 & 71.2                 & 79.2                 & 57.4                 & 64.7                 & 75.5                 & 59.6                 & 77.8                 & 74.8                 & 65.7                 & 74.5                  & 70.4                  \\
APE                   & 60.7                 & 81.6                 & 72.5                 & 78.6                 & 58.3                 & 63.6                 & 76.1                 & 53.9                 & 75.2                 & 72.3                 & 63.6                 & 69.8                  & 68.9                  \\
CDAC                  & 61.9                 & 83.1                 & 72.7                 & 80.0                 & 59.3                 & 64.6                 & 75.9                 & 61.2                 & 78.5                 & 75.3                 & 64.5                 & 75.1                  & 71.0                  \\
DECOTA               & 56.0                 & 79.4                 & 71.3                 & 76.9                 & 48.8                 & 60.0                 & 68.5                 & 42.1                 & 72.6                 & 70.7                 & 60.3                 & 70.4                  & 64.8                  \\
MCL                   & 67.0                 & 85.5                 & \textbf{73.8}                 & 81.3                 & 61.1                 & 68.0                 & 79.5                 & 64.2                 & 81.2                 & 78.4                 & 68.5                 & 79.3                  & 74.0                  \\ 
\midrule
\textbf{ProMM}  & \textbf{67.5} & \textbf{86.1} & 73.7 & \textbf{81.9} & \textbf{61.4} & \textbf{69.3} & \textbf{79.7} & \textbf{64.5} & \textbf{81.7} & \textbf{79.0} & \textbf{69.1} & \textbf{80.5} & \textbf{74.6} \\ 
\midrule
\multicolumn{14}{c}{\textbf{3-shot}} \\ 
\midrule
S+T                   & 55.7                 & 80.8                 & 67.8                 & 73.1                 & 53.8                 & 63.5                 & 73.1                 & 54.0                 & 74.2                 & 68.3                 & 57.6                 & 72.3                  & 66.2                  \\
DANN                  & 57.3                 & 75.5                 & 65.2                 & 69.2                 & 51.8                 & 56.6                 & 68.3                 & 54.7                 & 73.8                 & 67.1                 & 55.1                 & 67.5                  & 63.5                  \\
ENT                   & 62.6                 & 85.7                 & 70.2                 & 79.9                 & 60.5                 & 63.9                 & 79.5                 & 61.3                 & 79.1                 & 76.4                 & 64.7                 & 79.1                  & 71.9                  \\
MME                   & 64.6                 & 85.5                 & 71.3                 & 80.1                 & 64.6                 & 65.5                 & 79.0                 & 63.6                 & 79.7                 & 76.6                 & 67.2                 & 79.3                  & 73.1                  \\
APE                   & 66.4                 & 86.2                 & 73.4                 & 82.0                 & 65.2                 & 66.1                 & 81.1                 & 63.9                 & 80.2                 & 76.8                 & 66.6                 & 79.9                  & 74.0                  \\
CDAC                & 67.8                 & 85.6                 & 72.2                 & 81.9                 & 67.0                 & 67.5                 & 80.3                 & 65.9                 & 80.6                 & 80.2                 & 67.4                 & 81.4                  & 74.2                  \\
CLDA                   & 66.0                 & 87.6                 & \textbf{76.7}                 & 82.2                 & 63.9                 & 72.4               & 81.4                 & 63.4                 & 81.3                 & 80.3                 & 70.5                 & 80.9                  & 75.5                  \\ 
DECOTA              & 70.4                 & 87.7                 & 74.0                 & 82.1                 & 68.0                 & 69.9                 & 81.8                 & 64.0                 & 80.5                 & 79.0                 & 68.0                 & 83.2                  & 75.7                  \\
MCL                   & 70.1                 & 88.1                 & 75.3                 & 83.0                 & 68.0                 & 69.9                 & \textbf{83.9}                 & 67.5                 & \textbf{82.4}                 & 81.6                 & 71.4                 & \textbf{84.3}                  & 77.1                  \\ 
\midrule
\textbf{ProMM}  & \textbf{71.0} & \textbf{88.6} & 75.8 & \textbf{83.8} & \textbf{68.9} & \textbf{72.5} & \textbf{83.9} & \textbf{67.8} & 82.2 & \textbf{82.3} & \textbf{72.1} & 84.1 & \textbf{77.8} \\ 
\bottomrule
\end{tabular}}
    \caption{Accuracy (\%) on \textit{Office-Home} under the settings of 1-shot and 3-shot using ResNet34 as the backbone network.}
    \label{office-home}
\end{table*}

\begin{table}[t]
    \centering
    \tabcolsep=4.3pt
    \renewcommand\arraystretch{0.5}\scalebox{0.9}{
    \begin{tabular}{c|ccc|cc}
        \toprule
        &$L_{\text {intra}}$  & $L_{\text {inter}}$ & $L_{\text {batch}}$ &1-shot&3-shot\\
        \midrule
\ding{172} &\XSolidBrush     & \XSolidBrush     &\XSolidBrush & 74.1 & 78.1        \\
\ding{173} &\Checkmark     & \XSolidBrush     &\XSolidBrush & 77.6 & 82.9        \\
\ding{174} &\XSolidBrush  &\Checkmark          &\XSolidBrush  & 76.2 & 81.4       \\
\ding{175} &\XSolidBrush & \XSolidBrush  & \Checkmark & 75.1 & 80.0       \\
\ding{176} &\XSolidBrush & \Checkmark          & \Checkmark   & 83.1 & 84.6      \\
\ding{177}  &\Checkmark & \XSolidBrush     & \Checkmark   & 82.6 & 85.2      \\
\ding{178} & \Checkmark & \Checkmark          &\XSolidBrush   & 83.9 & 86.0      \\
\ding{179} & \Checkmark & \Checkmark    & \Checkmark     & \textbf{87.6} & \textbf{88.4}    \\
        \bottomrule
    \end{tabular}}
    \caption{Ablation studies of ProMM’s different components. We report the MCA (\%) on \textit{VisDA2017} under the settings of 1-shot and 3-shot using a ResNet34 backbone.}
    \label{ab}
\end{table}

\begin{table}[t]
\centering
\renewcommand\arraystretch{0.6}\scalebox{0.9}{
\begin{tabular}{c|ccc|cc} 
\toprule
  & \begin{tabular}[c]{@{}c@{}}linear \\pred.\end{tabular} & \begin{tabular}[c]{@{}c@{}}proto. \\pred.\end{tabular} & \begin{tabular}[c]{@{}c@{}}update \\proto.\end{tabular} & 1-shot   & 3-shot    \\ 
\midrule
\ding{172} & \Checkmark     & \XSolidBrush     &\XSolidBrush & 84.6     & 85.8\\
\ding{173} & \XSolidBrush     & \Checkmark     &\Checkmark & 80.7     & 84.2\\
\ding{174} & \Checkmark     & \XSolidBrush     &\Checkmark & 85.8     & 86.7\\
\ding{175} & \Checkmark     & \Checkmark    &\XSolidBrush & 82.0     & 85.1\\
\ding{176} & \Checkmark     & \Checkmark     &\Checkmark & \textbf{87.6}     & \textbf{88.4}\\
        \bottomrule
    \end{tabular}}
    \caption{Ablation study on the effectiveness of prototype-based contrast and update of our ProMM on \textit{VisDA2017} under the settings of 1-shot and 3-shot using a ResNet34 backbone.}
    \label{abpro}
\end{table}

\subsection{Overall Framework and Training Objective}
To sum up, the overall training objectives of the framework are as follows: 
\begin{align}\label{allloss}
L_{\text {all}} & = L_{\text {base}}+\lambda_{\text {intra}}L_{\text {intra}}+\lambda_{\text {inter}}L_{\text {inter}}+\lambda_{\text {batch}} L_{\text {batch}},
\end{align}
where $\lambda_{\text{intra}}$, $\lambda_{\text{inter}}$ and $\lambda_{\text{batch}}$ are the hyper-parameters that balance different levels.
We train the model in our framework by employing the overall training loss described in Equation \ref{allloss}.
\section{Experiments}
\subsection{Experimental Setup}
\textbf{Datasets.}
We verified our ProMM framework on three popular SSDA datasets.
\textbf{DomainNet} is the latest large-scale multi-source domain adaptive dataset, with 6 domains and 345 categories \cite{peng2019moment}.
According to \cite{saito2019semi}, four fields (Real, Clipart, Painting, Sketch) and 126 categories were selected for SSDA evaluation.
\textbf{VisDA2017} consists of 150k synthetic images and 55k real images, including two domains and 12 categories \cite{visda2017}.
For each category of each dataset, we randomly select one or three labeled samples (1-shot or 3-shot) as labeled target samples.
\textbf{Office-Home} is also a mainstream domain adaptive dataset, including 4 domains (Real, Clipart, Art, Product) and 65 classes \cite{venkateswara2017deep}.
We follow the standard of most SSDA work \cite{saito2019semi}, and report the overall accuracy as an indicator of \textit{DomainNet} and \textit{Office-Home}, and report the average class accuracy (MCA) as an evaluation indicator of \textit{VisDA2017}.

\noindent\textbf{Implementation details.}\footnote{\href{https://bupt-ai-cz.github.io/ProMM/}{https://bupt-ai-cz.github.io/ProMM/}}
To ensure the fairness of the experiment, similar to \cite{saito2019semi,li2021cross}, we use ResNet34 \cite{he2016deep} pre-trained on Imagenet as the backbone of the model.
The settings of batch size, optimizer, feature size, and learning rate are also consistent with \cite{saito2019semi}.
Similar to \cite{yan2022multi}, RandomFlip and RandomCrop are used as weak image augment methods, and RandAugment \cite{cubuk2020randaugment} is used as strong augment methods. The momentum $\alpha$ used to update prototypes is set to 0.9, and the threshold $\tau_1$ is set to 0.95 and $\tau_2$ is set to 0.4 for \textit{VisDA2017}, 0.3 for \textit{DomainNet} and 0.1 for \textit{Office-Home}.
For the OT solver, we solve it through Sinkhorn-Knopp iteration \cite{fatras2021unbalanced}.
The temperature $T_1$ and $T_2$ in the similarity function are set to 0.05 and 0.1.
Balance hyparameter $\lambda_{\text{intra}}$ and $\lambda_{\text{inter}}$ is set to 1.
$\lambda_{\text{batch}}$ is set to 1 for \textit{DomainNet}, 1 for \textit{Office-Home}, and 0.1 for \textit{VisDA2017}.

\subsection{Analysis of Experimental Results}
In the experiments, we compare our ProMM with two baselines: S+T uses only source and labeled target data, and ENT \cite{grandvalet2004semi} uses entropy minimization for unlabeled samples; and several popular SSDA methods, i.e., DANN \cite{ganin2016domain}, MME \cite{saito2019semi}, BiAT \cite{jiang2020bidirectional}, APE \cite{kim2020attract}, Con$^2$DA \cite{perez20222}, CDAC \cite{li2021cross}, DECOTA \cite{yang2021deep}, CLDA \cite{singh2021clda}, ECACL \cite{li2021ecacl} and MCL \cite{yan2022multi}.

\noindent\textbf{DomainNet.}
As shown in Tab.~\ref{multi}, our proposed ProMM achieves 76.1\% and 77.4\% average accuracy and SOTA performance in 7 scenarios of 1-shot and 3-shot, respectively.

\noindent\textbf{VisDA2017.}
\textit{VisDA2017} demonstrates the validity of ProMM in Tab.~\ref{visda2017}, our ProMM achieves 87.6\% MCA in 1-shot and 88.4\% in 3-shot, which are superior to the SOTA methods.

\noindent\textbf{Office-Home.}
As shown in Tab.~\ref{office-home}, ProMM outperforms the existing SOTA methods in both 1-shot and 3-shot scenarios, with accuracy reaching 74.6\% and 77.8\%, respectively.

\begin{figure*}[t]
\centering

\subfigure[Iteration:100]{
\begin{minipage}[t]{0.245\linewidth}
\centering
\includegraphics[width=0.99\textwidth,height=0.75\textwidth]{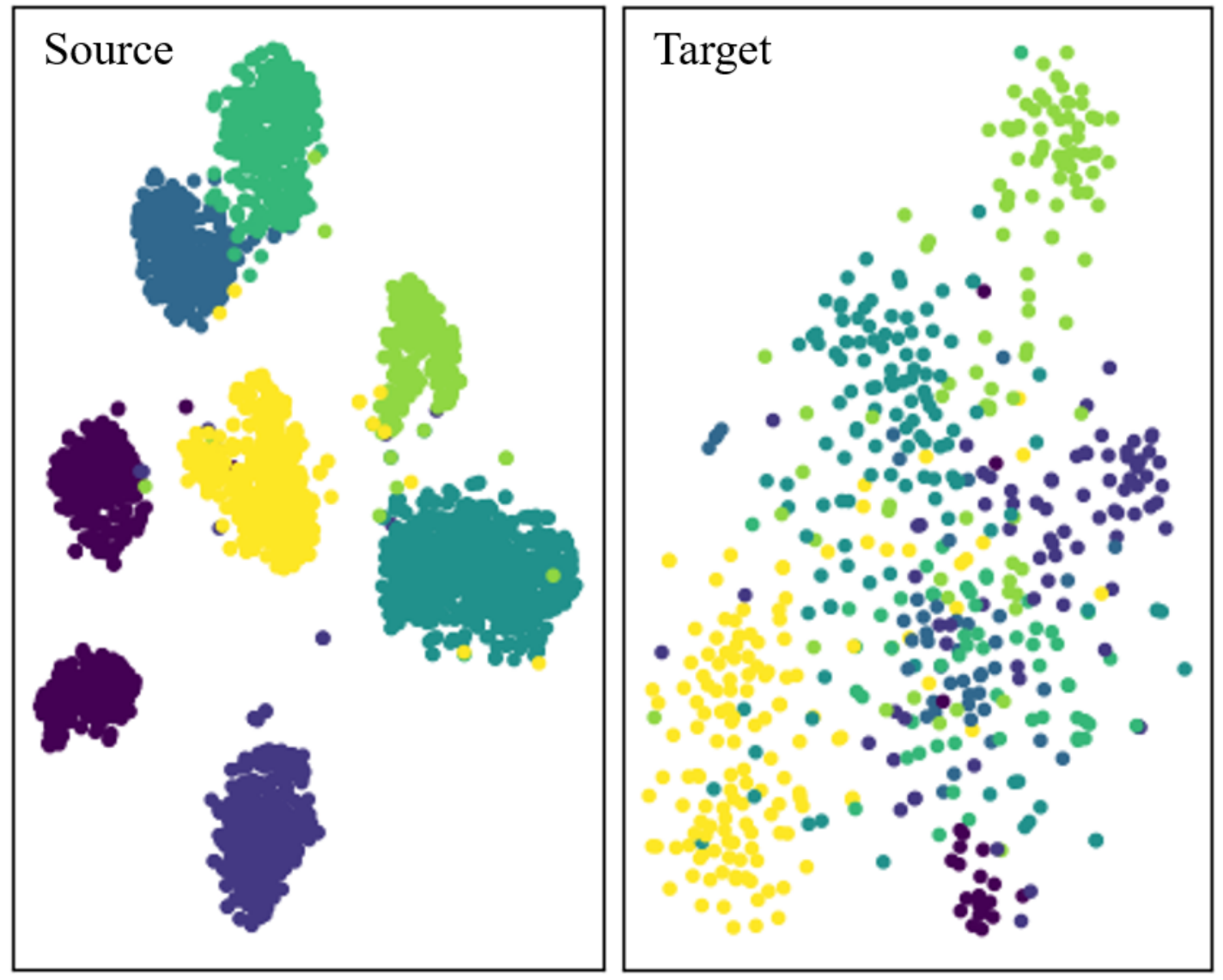}
\label{tsnea}
\end{minipage}%
}%
\subfigure[Iteration:1000]{
\begin{minipage}[t]{0.246\linewidth}
\centering
\includegraphics[width=0.99\textwidth,height=0.75\textwidth]{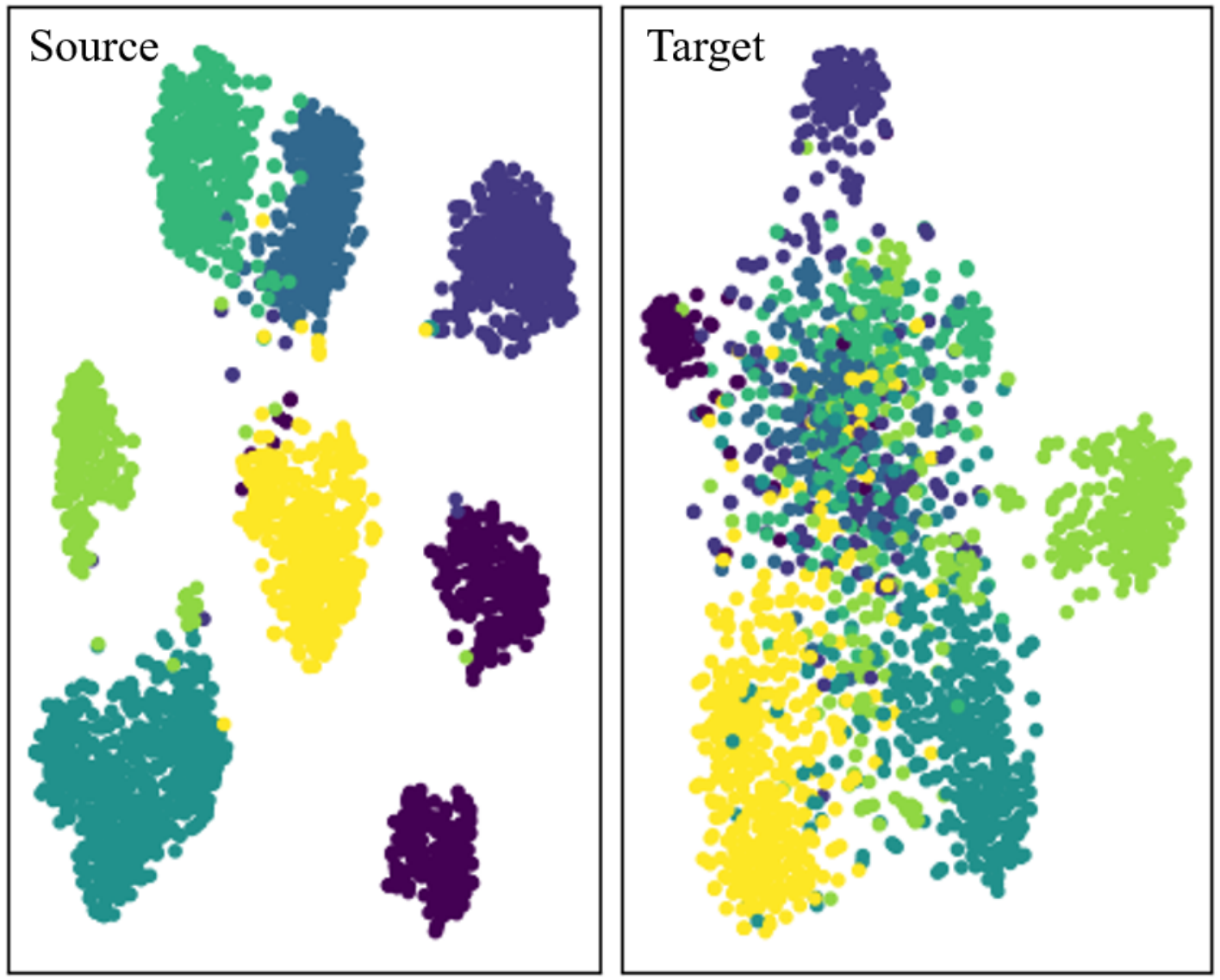}
\label{tsneb}
\end{minipage}%
}%
\subfigure[Iteration:5000]{
\begin{minipage}[t]{0.245\linewidth}
\centering
\includegraphics[width=0.99\textwidth,height=0.75\textwidth]{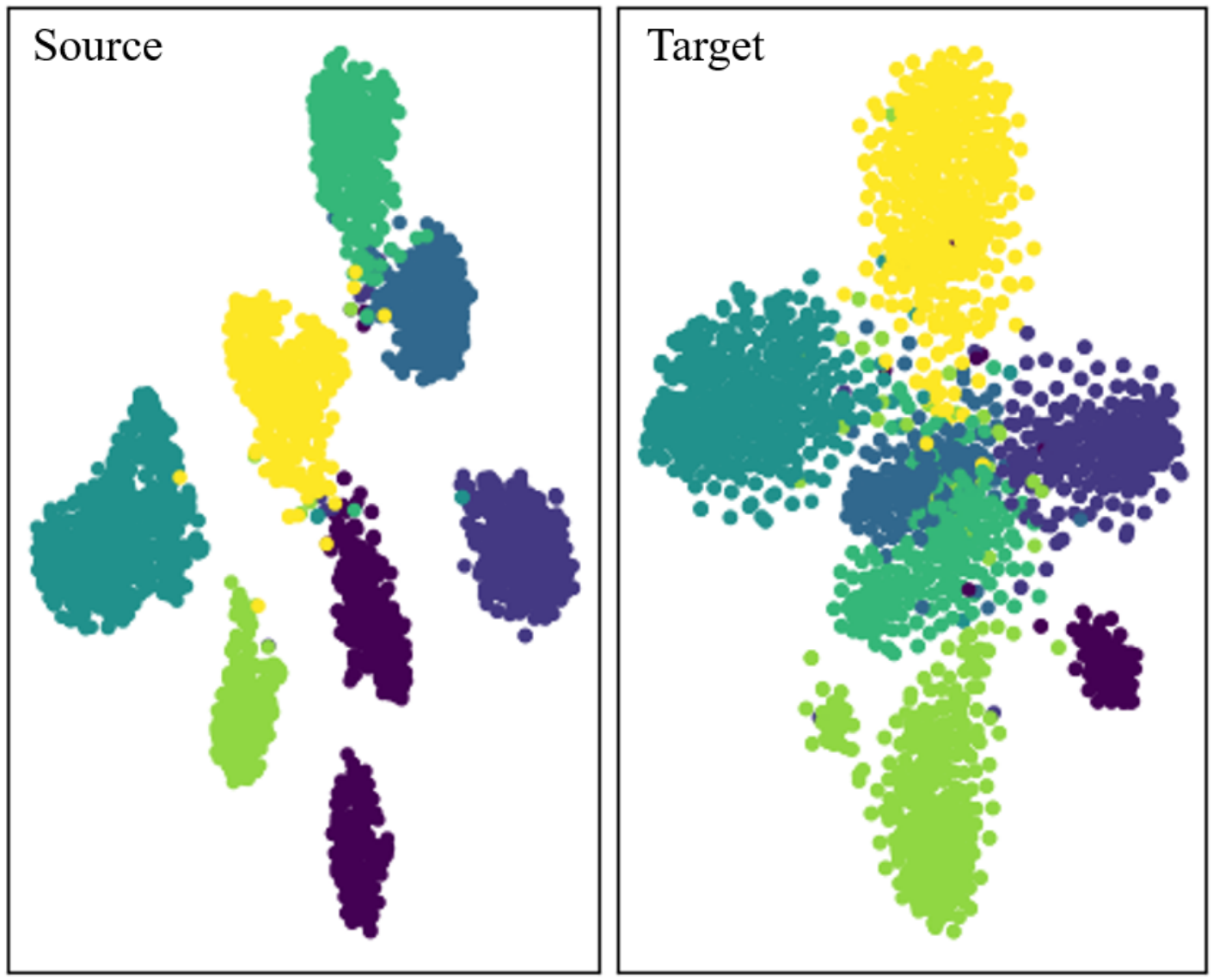}
\label{tsnec}
\end{minipage}
}%
\subfigure[Iteration:10000]{
\begin{minipage}[t]{0.245\linewidth}
\centering
\includegraphics[width=0.997\textwidth,height=0.75\textwidth]{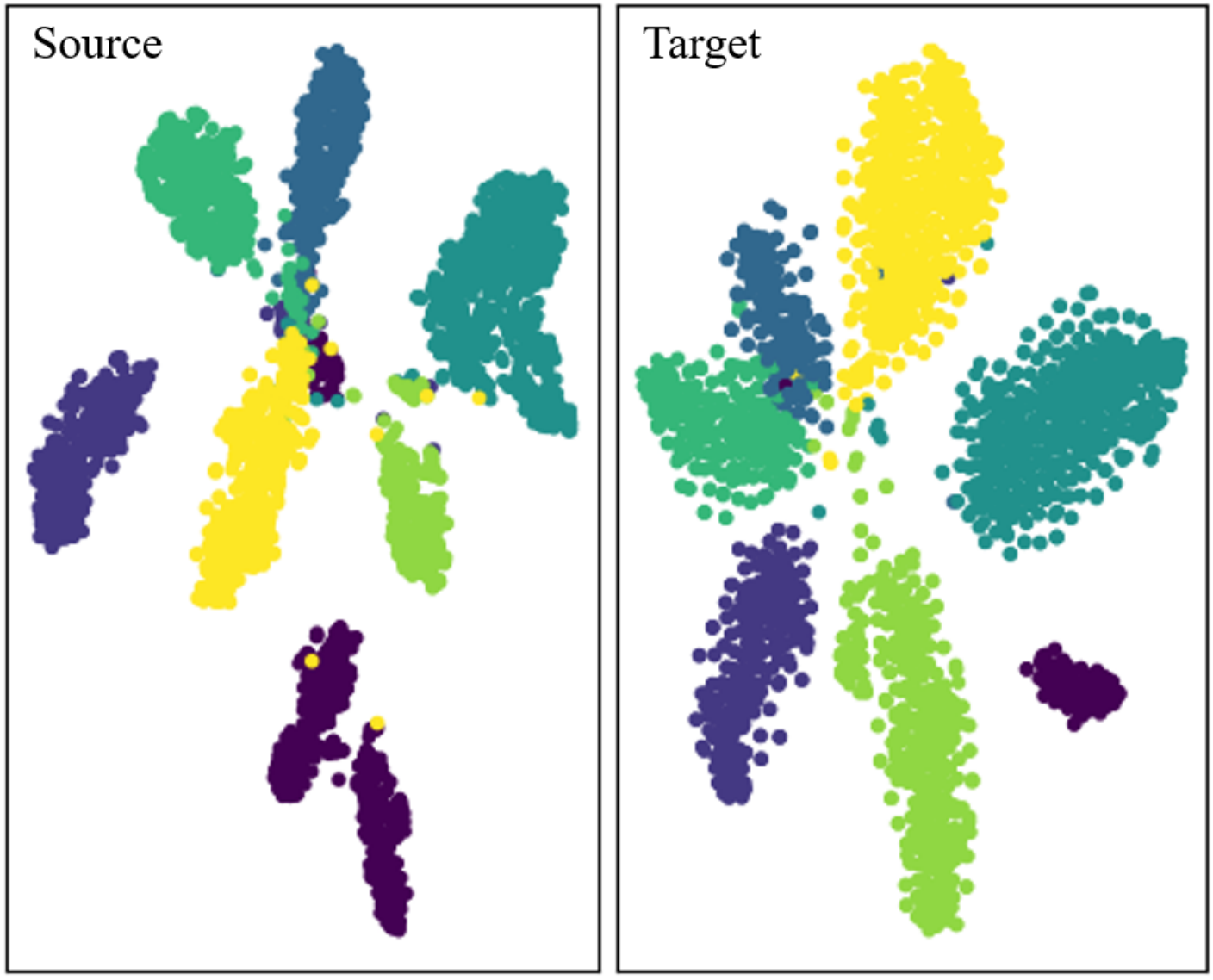}
\label{tsned}
\end{minipage}
}%

\centering
\caption{The t-SNE visualization of the feature alignment progress through our ProMM during the training time on \textit{\textit{DomainNet}} P$\rightarrow$R.}
\label{tsne}
\end{figure*}

\subsection{Additional Analysis}
\textbf{Ablation studies.}
We performed ablation experiments on \textit{\textit{VisDA2017}} at the settings of 1-shot and 3-shot, as shown in Tab. \ref{ab}.
Row \ding{173} \ding{174} \ding{175} show that each component can be significantly improved.
Row \ding{176} \ding{177} \ding{178} show that each combination still improves performance, indicating the universality of the proposed module.
Due to the integrity of our framework, one aspect of the lack of consideration may not be optimal in performance, but the best performance can be achieved when all components are activated.

\noindent\textbf{Prototype-based contrast and update.}
Tab. \ref{abpro} shows ablation studies of prototype-based classifier and updated procedure.
Row \ding{172} shows only linear prediction, and its performance degrades a lot, which proves that the knowledge learned only by using a linear classifier is limited.
The comparison with row \ding{174} proves that the model will benefit from the dynamic update of the prototype.
The global static prototype that only uses labeled data is not robust, so its generalization ability is limited.
Row \ding{173} shows only prototype-based predictions and dynamic updates during training.
Its performance is much lower, which means that without the help of a linear classifier, the prototype-based classifier cannot learn consistency under limited label data settings, so it cannot make full use of the additional knowledge brought by prototypes.
Row \ding{175} only adds the prototype-based classifier on row \ding{172}, but its performance does not increase but decreases, which indicates that biased prototypes will bring negative knowledge transfer.
Row \ding{176} achieves the optimal performance after integrating all methods, which shows that our method can complement knowledge from multi-level.

\begin{figure}[t]
\centering
\includegraphics[scale=0.21]{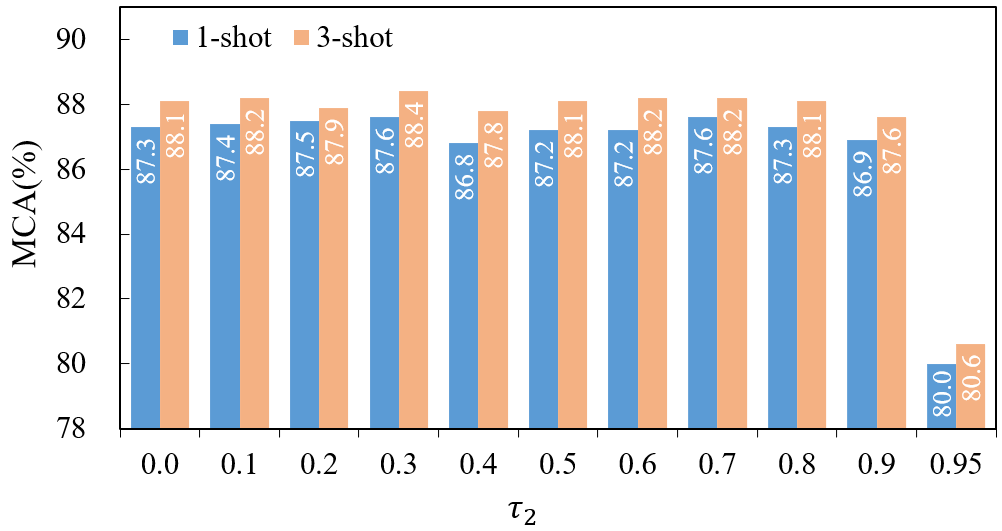}
\caption{Mean Class Accuracy (MCA)(\%) under different pseudo-label confidence thresholds on \textit{VisDA2017} for 1-shot and 3-shot, where $\tau_2=0.95$ means no intra-domain OT.}
\label{sensi}
\end{figure}

\begin{table}[t]
\centering
\renewcommand\arraystretch{0.8}
\tabcolsep=0.8pt\scalebox{0.9}{
\begin{tabular}{c|cccccc|cc} 
\toprule
\multirow{2}{*}{Method} & \multicolumn{2}{c}{\begin{tabular}[c]{@{}c@{}}Epoch\\100\end{tabular}} & \multicolumn{2}{c}{\begin{tabular}[c]{@{}c@{}}Epoch\\1000\end{tabular}} & \multicolumn{2}{c|}{\begin{tabular}[c]{@{}c@{}}Epoch\\10000\end{tabular}} & \multicolumn{2}{c}{Mean}  \\
                        & 1-shot & 3-shot                                                        & 1-shot & 3-shot                                                         & 1-shot & 3-shot                                                           & 1-shot & 3-shot           \\ 
\midrule
Linear                  &      \textbf{23.4}  &    36.0                                                           &   53.1     &     67.2                              &      65.6  &             76.5                                                     &    65.6    &           75.6       \\
Proto.                  &      18.8  &              \textbf{37.5}                                                 &     56.3   &           64.0                                                     &    65.6    &        74.9                                                          &    65.5    &          74.1        \\ 
\midrule
Our                     &       \textbf{23.4} &      \textbf{37.5}                                                         &    \textbf{57.3}    &    \textbf{68.1}                                                          &    \textbf{68.1}    &                   \textbf{78.1}                                               &   \textbf{66.9}     &          \textbf{76.8}        \\
\bottomrule
\end{tabular}}
\caption{The label accuracy (\%) of different pseudo label methods under the 1-shot and 3-shot settings of \textit{\textit{DomainNet}} C$\rightarrow$S. \textit{Linear} represents the linear classifier probability, \textit{Proto.} represents the prototype similarity, and \textit{Our} represents our intra-domain OT strategy.}
\label{pseudoacc}
\end{table}

\noindent\textbf{Intra-domain OT pseudo-labels.}
Tab. \ref{pseudoacc} shows the pseudo-label accuracy of different methods.
By making the target domain more compact, our strategy gives more accurate pseudo-labels at different training stages than simple methods only using maximum linear classifier prediction (\textit{Linear}) or maximum prototype similarity (\textit{Proto.}) under the same confidence threshold $\tau_2$ in Equation \ref{label}.

\noindent\textbf{Convergence analysis.} 
To further analyze the convergence of our method, we describe the feature t-SNE \cite{van2008visualizing} visualization of source and target domains during different training times in Figure~\ref{tsne}. 
We randomly selected seven categories of the feature on \textit{DomainNet} R$\rightarrow$S for clearer visualization. 
Figure \ref{tsnea} clearly shows the initial domain difference between the source and the target domain.
Early feature descriptions often show many misaligned source and target clusters, so the model initially performs poorly in the target domain. 
As the training progresses, it can be seen from Figure \ref{tsneb} and Figure \ref{tsnec} that our method converges and aggregates the feature of the target domain.
Accumulated good target feature is obtained and it proves that our method can obtain a compact target domain distribution, as shown in Figure \ref{tsned}.

\noindent\textbf{Parameter sensitivity.}
We will use the pseudo-labels threshold 
 $\tau_ 2$ of intra-domain OT strategy in Equation \ref{label} as a variable, analyze changes in MCA with \textit{\textit{VisDA2017}} under the settings of 1-shot and 3-shot, and experimental results support our argument.
As shown in Figure \ref{sensi}, with the change of the threshold, MCA floats within 1\%, which demonstrates the robustness of the strategy.
When this strategy is not used, i.e. $\tau_2=0.95$, MCA decreased significantly.
This result is reasonable and shows that this strategy can help the model make use of unlabeled target samples.

\noindent\textbf{Impact of the number of target labels. }
We studied the performance impact of different target label numbers on \textit{Domain} R$\rightarrow$S. 
From Figure \ref{number}, all methods can improve performance by labeling more target samples. 
By contrast, our ProMM always achieves optimal performance in all cases. 
This confirms that our method can make use of a flexible number of labeled target samples to help model knowledge transfer.

\noindent\textbf{Discussion of Difference from MCL.}
We have supplemented a table (cf. Tab.\ref{visda2017}) with explanations of the difference from MCL.
Our paper is completely different from MCL, including inter- and intra-domain, and we have better results.
MCL solves SSDA from the perspective of consistency learning in semi-supervised learning, without considering how to better utilize labeled target data.
We use labeled target data to generate prototypes that are more suitable for SSDA.
We considered not only consistency but also other aspects based on target prototypes.
We design a novel intra-domain pseudo-label strategy, inter-domain prototype alignment, and a new batch-wise dual contrast learning using target prototypes that MCL does not have.
These are all things that MCL without using labeled target data cannot do, which are not reflected in existing work.
\begin{table}[ht]
\centering
\setlength{\tabcolsep}{6.5mm}{
\renewcommand{\arraystretch}{1.5}
\scalebox{0.64}{
\begin{tabular}{c|c|c}
\toprule
Method  & MCL  & ProMM (Ours) \\
\hline
Motivation     & \thead{Consistent learning inspired \\by semi-supervised learning}          & \thead{Prototype-based learning \\using target labels}        \\
\hline
\multirow{3}*{\thead{Different Methods}}  & \multirow{3}*{\thead{Multi-level\\Consistency Regularization}} & \thead{Pseudo-label Strategy} \\
\cline{3-3}
~  & ~ & \thead{Prototype Alignment} \\
\cline{3-3}
~  & ~ & \thead{Dual Consistency} \\
\bottomrule
\end{tabular}}}
\caption{Difference from MCL.}
\label{visda2017}
\end{table}

\begin{figure}[t]
\centering
\includegraphics[scale=0.197]{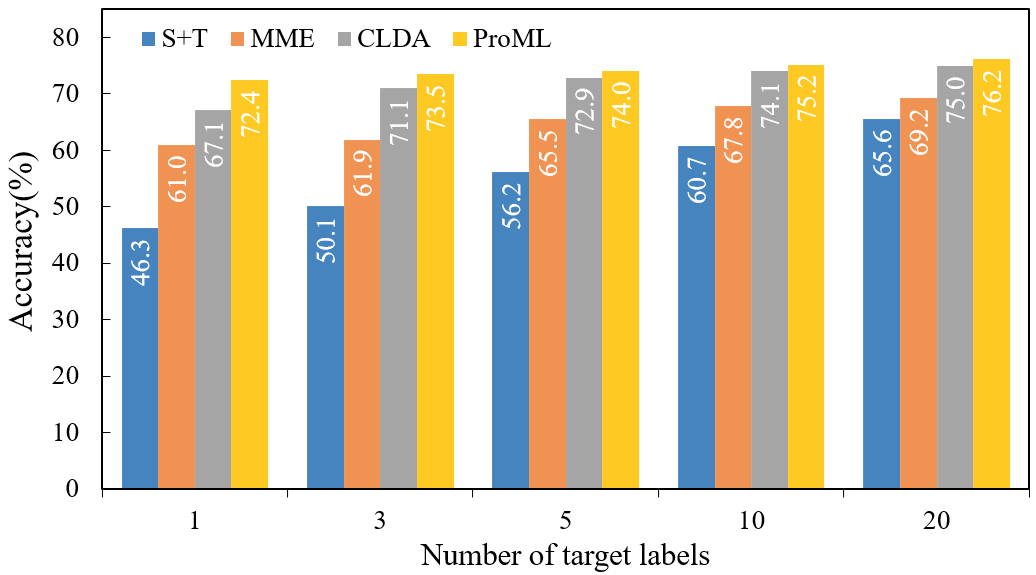}
\caption{Accuracy with different numbers of labeled samples per class in target domain on \textit{DomainNet} R$\rightarrow$S.}
\label{number}
\end{figure}

\section{Conclusion}
In this paper, we propose a novel Prototype Multiple-perspective Mining (ProMM) framework to make better use of target samples.
ProMM leverages prototypes constructed from target samples at three levels, (i) intra-domain level, aligns labeled and unlabeled sample distributions within the target domain using pseudo-label aggregation help models based on optimal transport, (ii) inter-domain level, it aligns source and target domain, and (iii) batch level, it learns compact classes and partition clusters from two dual perspectives.
Extensive experimental studies have demonstrated its advantages.

\section*{Acknowledgement}
This work was supported by the National Key R\&D Program of China (2021ZD0109802), by the National Natural Science Foundation of China (81972248).

\bibliographystyle{named}
\bibliography{ijcai23}

\end{document}